\documentclass[10pt,journal,compsoc]{IEEEtran}

\ifCLASSOPTIONcompsoc
  \usepackage[nocompress]{cite}
\else
  \usepackage{cite}
\fi
\usepackage{graphicx}
\usepackage{amsmath}
\usepackage{amssymb}
\usepackage{epsfig}
\usepackage{times}
\usepackage{multirow}
\usepackage{ragged2e}
\usepackage[justification=centering]{caption}
\newcommand\norm[1]{\left\lVert#1\right\rVert}

\ifCLASSINFOpdf
\else
\fi
\usepackage{url}

\begin{document}
%
\title{CAN: Composite Appearance Network for Person Tracking and How to Model Errors in a Tracking System}
%
%
%
%

\author{Neeti Narayan,
        Nishant Sankaran,
        Srirangaraj Setlur,~\IEEEmembership{Senior Member IEEE,}
        and~Venu Govindaraju,~\IEEEmembership{Life~Fellow,~IEEE}
\IEEEcompsocitemizethanks{\IEEEcompsocthanksitem All authors are with the Department
of Computer Science and Engineering, University at Buffalo, NY, 14260.\protect\\
E-mail: neetinar@buffalo.edu
}
}

\IEEEtitleabstractindextext{%
\justify
\begin{abstract}
Tracking multiple people across multiple cameras is an open problem. It is typically divided into two tasks: (i) single-camera tracking (SCT) - identify trajectories in the same scene, and (ii) inter-camera tracking (ICT) - identify trajectories across cameras for real surveillance scenes. Many methods cater to SCT, while ICT still remains a challenge. In this paper, we propose a tracking method which uses motion cues and a feature aggregation network for template-based person re-identification by incorporating metadata such as person bounding box and camera information. We present a feature aggregation architecture called Composite Appearance Network (CAN) to address the above problem. The key structure of this architecture is called \textit{EvalNet} that pays attention to each feature vector and learns to weight them based on gradients it receives for the overall template for optimal re-identification performance. We demonstrate the efficiency of our approach with experiments on the challenging multi-camera tracking dataset, DukeMTMC. We also survey existing tracking measures and present an online error metric called ``Inference Error'' (IE) that provides a better estimate of tracking/re-identification error, by treating SCT and ICT errors uniformly.
\end{abstract}

\begin{IEEEkeywords}
Person re-identification, Person tracking, Feature aggregation, Error metric
\end{IEEEkeywords}}

\maketitle

\IEEEdisplaynontitleabstractindextext

%
\IEEEpeerreviewmaketitle

\ifCLASSOPTIONcompsoc
\IEEEraisesectionheading{\section{Introduction}\label{sec:introduction}}
\else
\section{Introduction}
\label{sec:introduction}
\fi

%
%
%
%
\IEEEPARstart{P}{erson} tracking is one of the fundamental problems in the field of computer vision. Determining multi-camera person trajectories enable applications such as video surveillance, sports, security, behavior analysis and anomaly detection. With the increase in the amount of video data and the number of deployed
cameras in recent years, it becomes essential to have an automated
and reliable tracking system. Maintaining the correct trajectory of a person across cameras is difficult because of occlusion, illumination change, background clutter, and blind-spots due to non-overlapping cameras often placed far apart to reduce costs. Methods like \cite{kalal2012tracking} focus on target tracking using a low-level handcrafted feature. Although they achieve good tracking performance, they are still inefficient in solving the mentioned obstacles and are limited to scenarios where targets are known a-priori. Some model people entry and exit points on the ground or image plane explicitly \cite{cai2014exploring, gilbert2006tracking}. Few others \cite{berclaz2011multiple, zhang2015tracking} exploit data fusion methods from partially overlapping views or completely overlapping views as a pre-processing step.

\begin{figure}
\begin{center}
\centerline{\includegraphics[width=1\linewidth]{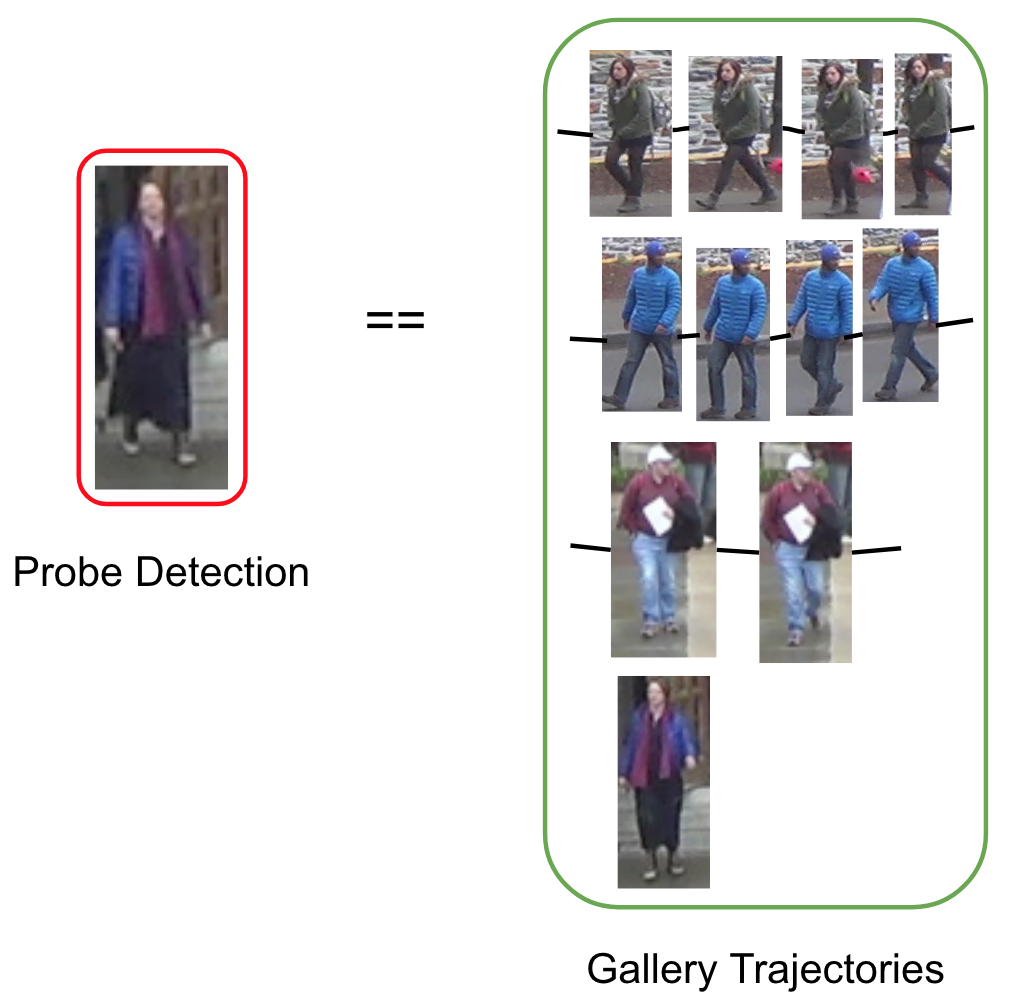}}
\end{center}
\vspace{-0.4cm}
  \caption{Example of unconstrained tracking. Given a probe image, the problem is to find its matching trajectory from gallery.}
\label{fig:can_outline}
\end{figure}

Person re-identification for the purpose of estimating the similarity of person images which deals with data comprised of only image pairs, one for the probe and one for the gallery, is relatively straightforward with the use of metric learning methods \cite{liao2015person, zhang2016learning} and feature representation learning \cite{kviatkovsky2013color}. However, in unconstrained tracking (Figure \ref{fig:can_outline}), person tracklet contains multiple images and therefore requires a way to fuse features/attributes to a single feature vector representative of the tracklet. Feature quality of person bounding box attributed to the contribution of identifying an individual is different or inconsistent at different time instances, therefore a straightforward feature aggregation is insufficient to result in optimal feature fusion. We design a spatio-temporal attention model, Composite Appearance Network (CAN), where the network looks at each individual feature vector of a trajectory in the gallery and predicts how important it is to be a part of the final representative feature vector by incorporating metadata such as person bounding box and camera information. We also discuss the Inference Error (IE) evaluation criterion to measure the performance of online person tracking. 



The main contributions of this paper include:
\begin{itemize}
\item Finding the optimization function that best exploits the variances between the appearance features (extracted from a CNN or any other embedding system) using additional metadata. This would result in optimal aggregation weights for pooling.  
\item We experiment CAN model on a large-scale multi-person multi-camera tracking dataset and obtain improvements over baseline.
\item Study the performance measures of a tracking system and demonstrate how they are different when evaluating/ characterizing online and real-time tracking performance. We formally introduce the ``Inference Error'' metric and emphasize on de-duplication.
\end{itemize}

The paper is organized as follows: In the following section, background and related work in person tracking are discussed. Also, existing feature aggregation methods are elaborated. Before describing CAN model, the different error metrics proposed for tracking are first reviewed and use cases of IE are defined in Section \ref{sec:measures}. Section \ref{sec:framework} provides details of our feature aggregation network to learn representative ID-features, and describes the different parts of our tracking framework. Finally, experimental details and evaluation results are presented in Section \ref{sec:experiments}.
\section{Related Work}

\subsection{Person Tracking}
Visual tracking using Convolutional Neural Networks (CNN) has gained popularity, attributed to CNNs performance in representing visual data \cite{nam2016learning, wang2015transferring}. Existing visual tracking systems focus on finding the location of the target object. In \cite{nam2016learning}, a tracking-by-detection approach based on a CNN trained in a multi-domain learning framework is proposed. The attention is on improving the ability to distinguish the target and the background. To deal with issues resulting from the requirement for large amount of labeled tracking sequences, and inefficient search algorithms, such as sliding window or candidate sampling, a new tracker based on reinforcement learning is proposed in \cite{yooaction}. An action-decision network (ADNet) is designed to generate actions to find the location and the size of the target object in a new frame. The framework solves the insufficient data problem. However, results reported are based on metrics like center location error and bounding box overlap ratio.


In \cite{jin2019multi}, a unified framework for multi-person pose estimation and tracking with spatial and temporal embeddings is proposed. The spatial module detects body parts and performs part-level data association in a single frame. The temporal module groups human instances in consecutive frames to track people across time. Multi-person pose estimation in videos has been studied also in \cite{iqbal2016pose}. They jointly perform pose estimation and tracking, but is limited to single camera videos. The problem of person re-identification is not addressed, rather they aim to solve the association of each person across the video as long as the person does not disappear. For multi-camera object tracking, input detections can be regarded as a graph and weight edges between nodes (detections) based on similarity \cite{chen2017equalized}. When merging tracklets, trackers are dependent on the feature extractor/ representation technique to avoid high false negatives. For example, some trackers adopt LOMO \cite{liao2015person} appearance features and hankel matrix based IHTLS \cite{dicle2013way} algorithm for motion feature. A hard mining scheme and an adaptive weighted triplet loss is proposed in \cite{ristani2018features} to learn person appearance features from individual detections.


In \cite{sadeghian2017tracking}, an online method for tracking is proposed by using Long Short-Term Memory (LSTM) networks and multiple cues such as appearance, motion, and interaction. However, this solution is for single camera tracking and there is no proof of how well the system scales for multi-camera environment. In \cite{narayan2017person}, authors address the multi-person multi-camera tracking problem by reformulating it as a pure re-identification task. The objective is to minimize misassociations at every timestep. For this reason, the authors propose a new evaluation metric called the ``Inference Error'' (IE). The method, however, does not encode long-term spatial and temporal dependencies which are critical for correcting data association errors or recovering from occluded state. We discuss in detail the advantages of using IE and how it adapts to both online and offline tracking systems. An online tracking method, which is an extension of \cite{narayan2017person}, is presented in \cite{narayan2018re}. LSTM-based space-time tracker is developed using history of location and visual features to learn from past association errors and make a better association in the future. But, the model is not trained end-to-end. 

\subsection{Feature Aggregation}
In \cite{babenko2015aggregating}, local features such as Fisher vectors are compared with deep features for aggregation. Traditional hand-crafted features have different distributions of pairwise similarities, which requires careful evaluation of aggregation methods. In \cite{zhu2018online}, a temporal attention network (TAN) is proposed for multi-object tracking. It adaptively allocates different degrees of attention to different observations and filters out unreliable samples in the trajectory. Very recently, feature pooling has been explored to assess the quality of facial images in a set and sometimes it relies on having to carefully define ``weighting functions'' to produce intelligent weights. Neural Aggregation Networks (NAN) \cite{yang2017neural} fuse face features with a set of content adaptive weights using a cascaded attention mechanism to produce a compact representation. An online unsupervised method for face identity learning from unconstrained video streams is proposed in \cite{Pernici_2018_CVPR} by coupling CNN based face detection and descriptors with a memory based learning mechanism. 

In video-based person re-identification works, researchers explore temporal information related to person motion. In \cite{xu2017jointly}, features are extracted by combining color, optical flow information, recurrent layers and temporal pooling in a Siamese network architecture. Quality Aware Network (QAN) \cite{liu2017quality} learns the concept of quality for each sample in a set using a quality generation unit and feature generation part for set-to-set recognition, which learns the metric between two image sets. However, most of these methods learn each trajectory's representation separately and invariably rely on the individual features of person detection sequences, without considering the influence of the trajectory being associated with. Also, many use Recurrent Neural Network (RNN) to handle sequential data (input and output). It is possible to avoid them by borrowing their attention mechanism/differentiable memory handling into a simple feature aggregation framework. Hence, in order to draw different attention when associating different pairs of trajectories, we propose to use orthogonal attributes such as metadata, that are learnt using different target objectives.

The framework proposed in this paper is inspired from \cite{icb17}. The authors propose set based feature aggregation network (FAN) for the face verification problem. By generating representative template features using metadata like yaw, pitch and face size, their system could outperform traditional pooling approaches. Instead of building a siamese network, we focus on learning the pooled feature for gallery template (trajectory) while coupling the probe metadata beside gallery metadata. This way, we derive each trajectory's representation by considering the impact or influence of the probe detection in the context of data association.

\section{Performance Measures}
\label{sec:measures}
In this section, we survey the many existing tracking measures and discuss their use cases. 
There are two parts crucial to a multi-person multi-camera tracking (MPMCT) system: (i) SCT: single-camera (within-camera) tracking module and (ii) ICT: inter-camera (handover) tracking (ICT) module. Few metrics evaluate the two separately, some handle both simultaneously. Also, based on the application, measures can be classified into (i) event-based: count how often a tracker makes mistakes and determine where and why mistakes occur, and (ii) identity-based: evaluate how well computed identities conform to true identities.

\begin{figure*}
\centering

\centerline{\includegraphics[width=1.0\linewidth]{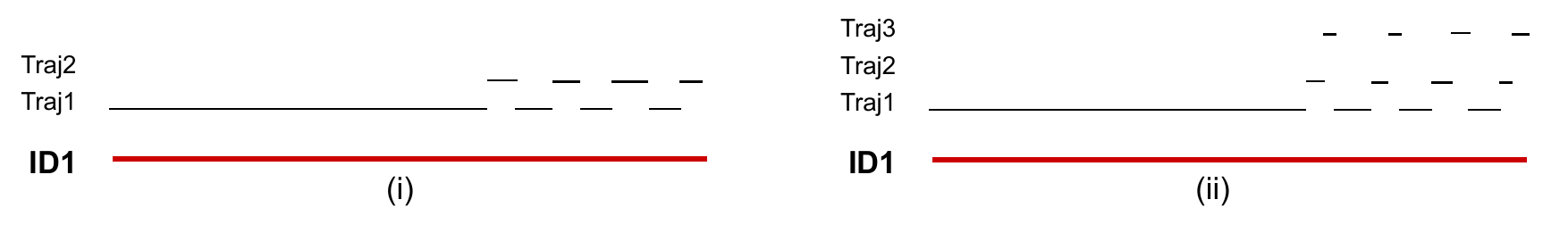}}
  \caption{Illustration of duplication effect. For one true identity $ID1$, (i) two trajectories are created and (ii) three trajectories are created in gallery due to misassociations. The two cases should be penalized differently to reduce fragmentations and control growth of dynamically changing gallery set.}
\label{fig:abc}
\end{figure*}
\subsection{Multi Object Tracking Accuracy (MOTA)}


MOTA (Multiple Object Tracking Accuracy) is typically used to evaluate single camera, multiple person tracking performance. It is defined as:
\begin{equation}
MOTA = 1 - \dfrac{FN+FP+Fragmentation}{D}
\end{equation}
where FN is the number of false negatives i.e. true targets missed by the tracker, FP is the number of false positives i.e. an association when there is none in the ground truth, Fragmentation is the ID switches, and D is the number of detections. However, MOTA penalizes detection errors, under-reports across-camera errors \cite{chen2017equalized} and can be negative due to false positives.  

\subsection{Identity-based Metrics}
We evaluate our framework on DukeMTMC using measures proposed in \cite{ristani2016performance}: Identification F-measure ($IDF1$), Identification Precision ($IDP$) and Identification Recall ($IDR$).  
Ristani et al. \cite{ristani2016performance} propose to measure performance by how long a tracker correctly identifies targets, as some users may be more interested in how well they can determine who is where at all times. Few critical aspects such as where or why mismatches occur are disregarded. $IDP$ is the fraction of computed detections that are correctly identified. $IDR$ is the fraction of ground truth detections that are correctly identified. $IDF_1$ is the ratio of correctly identified detections over the average number of ground-truth and computed detections. 

\subsection{Multi-camera Object Tracking Accuracy (MCTA)}
Another existing multi-camera evaluation metric is multi-camera object tracking accuracy (MCTA) \cite{chen2017equalized}. Here, both SCT and ICT are considered equally important in the final performance measurement. So, all aspects of system performance are condensed into one measure and defined as:
\begin{equation}
MCTA = F_1 * \bigg(1-\dfrac{M^s}{{TP}^s}\bigg)\bigg(1-\dfrac{M^i}{{TP}^i}\bigg)
\end{equation}
where the first term $F_1$-score measures the detection power (harmonic mean of precision and recall). The second term penalizes within-camera mismatches ($M^s$ = ID-switches) and normalized using true positive detections (${TP}^s$). The final term penalizes inter-camera mismatches ($M^i$) normalized by true inter-camera detections (${TP}^i$). MCTA factors single-camera mismatches, inter-camera mismatches, false-positives and false negatives through a product instead of adding them. In addition, error in each term is multiplied by the product of other two terms which might drastically change the performance measure.

\subsection{Inference Error}
Traditional biometric measures \cite{mansfield2002best} such as FMR (False Match Rate), FNMR (False Non-match Rate) assume that the occurrence of an error is a static event which cannot impact future associations. However, in a tracking by re-identification system, the reference gallery is dynamically evolving, as new tracks are created (following ``no association'' outcomes) or existing tracks are updated (following ``association'' outcomes).

For this reason, we formally introduce the ``Inference Error'' measure for tracking by continuous re-identification evaluation and is defined as:
\begin{equation}
E = \dfrac{1}{T}\sum_{t=1}^{T} \dfrac{M_t}{D_t}
\end{equation}
where $M_t$ is the number of misassociations at time $t$ and $D_t$ is the number of current detections.
The inference error is a real-time evaluation paradigm which measures how often a target is incorrectly associated. We normalize it by the number of detections. It handles multiple simultaneous observations in any given time and also handles identities reappearing in the same camera or in different cameras. 

The main intuition behind counting the number of misassociations at every timestep is to reduce creating duplicate trajectories for an identity. Consider one true identity $ID1$ as illustrated in Figure \ref{fig:abc}. Let time be represented in the horizontal direction. In case (i), a tracker mistakenly identifies two trajectories ($Traj1$ and $Traj2$) for the same person. While in case (ii), three trajectories ($Traj1$, $Traj2$ and $Traj3$) are mistakenly identified for the same person. If $Traj1$ covers $80\%$ of $ID1$'s path, identification measure would charge $20\%$ of the length of $ID1$ to each of the two cases. However, inference error would assign a higher penalty to case (ii) owing to multiple fragmentations. Given a real-time large-scale tracking scenario with multiple simultaneous observations across multiple cameras, system designers and security professionals will want a tracker that maintains the identity of a person and effectively updates the reference gallery without growing exponentially. So, identifying where and when trajectories are broken is essential for de-duplication. De-duplication is analogous to tracking framework, with the advantage of describing duplication errors (i.e., instances of incorrect ``non-match'' matching outcomes resulting in duplicate data, that would otherwise maintain the same identity) better.

\section{Proposed Tracking Framework}
\label{sec:framework}
The input to the multi-person multi-camera tracking (MPMCT) system is a set of videos from each camera. The output is a set of trajectories across all cameras. We do not take the end-to-end training approach to solve MPMCT. Training a network that is responsible for both detection and tracking, with a single loss is difficult and one could fail to learn weights in the early layers of such a deep model. Also, back-propagating the loss to find optimal parameters that predict true trajectories is a combinatorial problem and hence, an expensive solution. Here, we take the tracking solution a step further by proposing a simple feature aggregation approach that can well distinguish identical pair of features from non-identical ones to maintain a margin between within-identity and between-identity distances. 


\subsection{Preliminary Appearance Cues}
Given the input person detections in frame-level in a single camera or across multiple cameras over a time period $T$, we extract the appearance feature maps set using DenseNet  \cite{huang2017densely}. Dense Convolutional Neural Network (DenseNet) connects each layer to every other layer in a feed-forward fashion. The $l^{th}$ layer receives the feature maps of all preceding layers, $z_0, z_1,...,z_{l-1}$, as input. Thus, the feature map in the $l^{th}$ layer is given by:
\begin{equation}
z_l = U_l([z_0,z_1,...,z_{l-1}])
\end{equation}
where $[z_0,z_1,...,z_{l-1}]$ is the concatenation of feature maps of previous layers and $U_l$ can be considered as a composite function of operations such as batch normalization \cite{ioffe2015batch}, rectified linear units (ReLU) \cite{glorot2011deep}, etc. Concatenating feature maps in different layers results in lot of variation in the input of subsequent layers and improves efficiency. The other advantages of using DenseNet include eliminating vanishing-gradient problem, strengthening feature propagation, and reduction in the number of parameters.

We train DenseNet on $25\%$ of the training set (called trainval) of DukeMTMC dataset. Features of length $1024$ are extracted from the last dense layer. The DenseNet used has $4$ blocks with a depth of $121$, compression/reduction of $0.5$ and growth rate of $32$. We run for $15$ epochs with a learning rate starting at $0.1$ and we train using stochastic gradient descent with batch size $96$ and momentum 0.9. As the number of remaining epochs halves, we drop learning rate by a factor of 10 and drop by a factor of 2 at epoch $14$.

\begin{figure}
\begin{center}
\centerline{\includegraphics[width=1.0\linewidth]{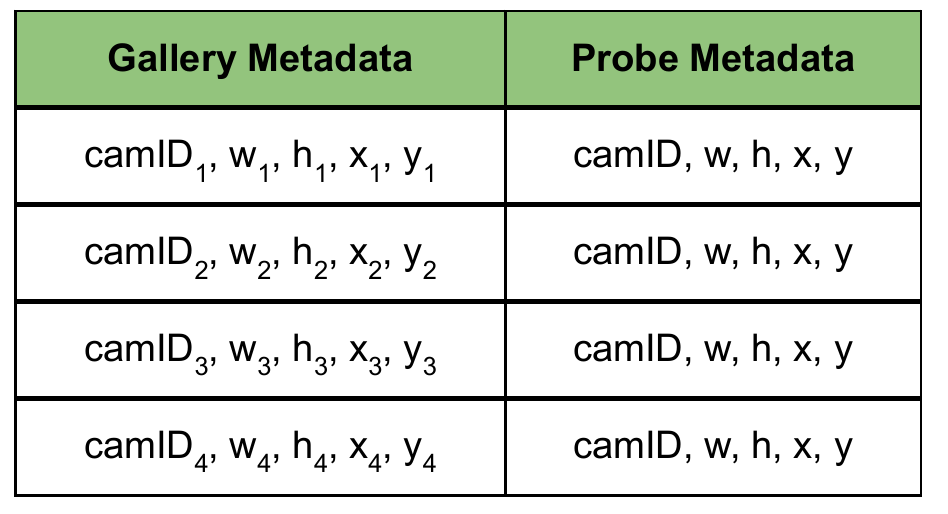}}
\end{center}
\vspace{-0.6cm}
  \caption{Illustration of gallery metadata list. Each row corresponds to a vector of length $10$.}
\label{fig:metadata}
\end{figure}

\subsection{CAN Architecture}
\begin{figure*}
\begin{center}
\centerline{\includegraphics[width=.9\linewidth, height=7.5cm]{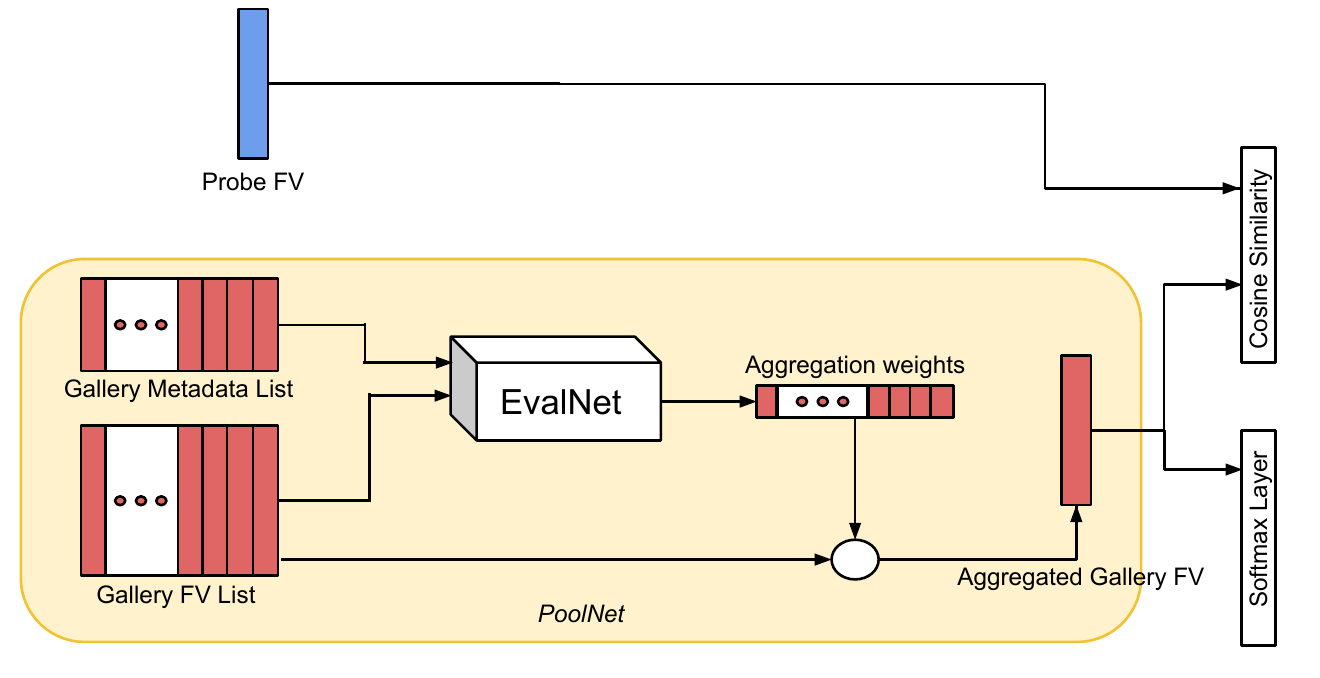}}
\end{center}
\vspace{-0.5cm}
  \caption{Composite Appearance Network architecture. EvalNet is a fully connected network that learns weights for producing the optimal aggregated feature vector.}
\label{fig:overview}
\end{figure*}

The Composite Appearance Network (CAN) is shown in Figure \ref{fig:overview}. The main component of CAN is EvalNet which evaluates the quality of every feature vector (FV) as being a part of the template (tracklet) by looking at appearance attribute and metadata simultaneously, and outputs a weight accordingly denoting the ``importance'' of that FV. 

The terms ``gallery'' and ``probe'' in the tracking scenario apply to tracklets identified dynamically as time progresses and a new detection to be associated with an existing tracklet respectively. The gallery FV list is one of the inputs to the network. It is comprised of a list of appearance attributes corresponding to a trajectory in the gallery. Gallery metadata list is the other input to the network. Metadata for person detections include bounding box information $(w, h, x, y) $ as well as other external details such as camera number $(camID)$. The probe metadata is concatenated with gallery metadata, which in-turn constitutes the gallery metadata list as illustrated in Figure \ref{fig:metadata}. Each vector in the list is of length $10$. In this way, the probe's influence is also taken into account for learning weights for aggregation. Probe FV is the probe detection's appearance attribute. Consider $n$ detections in a trajectory and corresponding appearance attributes of length $1024$. Then, the gallery FV list is of shape $(n\times1024)$, probe FV has shape $(1\times1024)$ and metadata list is $(n\times10)$.

EvalNet is a Fully Connected Network (FCN) with $4$ fully connected layers, batch normalization and ReLU as the activation function. This FCN block produces weights $e_k$ such that $\sum_{k=1}^{n} e_k = 1$, where $k$ is the $k^{th}$ feature vector and $n$ is the number of detections belonging to a tracklet. These weight predictions are then applied on the corresponding appearance features to obtain the aggregated appearance feature for every tracklet. The similarity between probe and aggregated FV is obtained using cosine similarity. The cosine similarity loss layer and softmax loss layer are optimized against the match label and class label respectively. During real-time tracking, we do not use entire CAN and instead, use only EvalNet to obtain aggregated FV given appearance attributes and metadata information for every tracklet.

\subsection{CAN Training}
Let $g_k$ and $V_k$ be the $k^{th}$ gallery FV and corresponding metadata vector, respectively, in a trajectory. If $F$ denotes the final composite or aggregated appearance FV, it can be defined as:
\begin{equation}
F = \sum_{k=1}^{n} H_{\theta}([g_k, V_k]) g_k
\end{equation}
where $H_{\theta}$ is the EvalNet function on gallery FV and metadata, parameterized by $\theta$, predicting a weight that evaluates the relative importance of the feature vector using accompanying metadata. $n$ is the number of feature vectors/ detections in the trajectory. We use orthogonal information such as probe metadata and gallery metadata to yield additional context that can help discriminate appearance features of an identity.

If $P$ is the probe FV, then the mean-squared loss function is defined as:
\begin{equation}
L_{mse}^{(i,j)} = \bigg(\dfrac{P_i.F_j}{\norm{P_i}^2\norm{F_j}^2} - Y(P_i,F_j)\bigg)^2
\end{equation}
and the categorical cross-entropy loss function is defined as:
\begin{equation}
L_{cce}^{(j)} = - \sum_{c=1}^{C}q_c^{(j)} \log\hat{q}_c^{(j)}
\end{equation}
where $Y(P, F) \in \{0,1\}$ is the ground-truth match label, $C$ is the number of classes (unique identities), $q_c$ is the ground-truth class label of the one-hot encoded vector $q$ and $\hat{q}$ is the normalized probability assigned to classes.
Given gallery-probe vectors, the objective of CAN is to find the optimal parameters $\theta$ that minimize the following cost function:
\begin{equation}
J(\theta) = \dfrac{1}{m_p}\dfrac{1}{m_g}\sum_{i=1}^{m_p}\sum_{j=1}^{m_g}L_{mse}^{(i,j)} + L_{cce}^{(j)}
\end{equation}
where $m_p$ is the number of examples in probe set and $m_g$ is the number of examples in gallery set.

\textbf{Batch processing:} Since each trajectory may be comprised of a variable number of person detections, we would require CAN to be trained on a single gallery trajectory in every iteration as making batches of trajectories would be difficult. However, we work around this problem by coupling an additional input - trajectory index $r$, that corresponds to the unique identity in the batch that each person (features and metadata) would be mapped to. This allows us to gather multiple sets of trajectories as a batch, enabling CAN to converge faster. The aggregated trajectory representation is computed using only the corresponding features as indexed by $r$.

\subsection{Tracking}
We include the motion property of each individual in our framework for within-camera tracking. This would help address one key challenge - handling noisy detections - because velocities of an individual can be non-linear due to occlusion or detections being noisy. Here, we propose to use a simple method to aggregate detections into single camera trajectories and further merge these trajectories using multi-camera CAN framework.

\subsubsection{Single-camera Motion and Appearance Correlation} \label{motionapp}

Single-camera trajectories are computed online by merging detections of adjacent frames in a sliding temporal window with an intersection of $50\%$. Let $i$ denote a previous detection and $j$ denote a current detection at timestep $t$. If two bounding boxes, ${BB}_i$ and ${BB}_j$, detected in neighboring windows are likely to belong to one target, the factors that effect this likelihood include: (i) appearance similarity and (ii) location closeness and is defined as $score_t^{(i,j)}$, computed as:

\begin{equation}
score_t^{(i,j)} = Sim(A_i,A_j) + f({BB}_i,{BB}_j)
\end{equation}
where $A_i, A_j$ are the appearance FVs, $Sim(A_i,A_j)$ is considered as the appearance similarity between two detections - given by cosine similarity formulation $Sim(A_i,A_j) = \dfrac{A_i.A_j}{\norm{A_i}^2\norm{A_j}^2}$, \\and $f({BB}_i,{BB}_j)$ is the Intersection over Union (IoU) function defined as:
\begin{equation}
f({BB}_i,{BB}_j) = \begin{cases}
    1, & \text{if $IoU({BB}_i, {BB}_j)\geq0.5$}.\\
    -\infty, & \text{otherwise}.
  \end{cases}
\end{equation}

Adjacent detections in a camera are merged by applying a greedy technique of selecting associations based on decreasing scores. The output of this step will include a set of single-camera trajectories $S$ from all cameras. The number of trajectories in each camera is not a constant and varies from one camera to another.

\subsubsection{Multi-camera}
Once we have associated detections to tracklets or trajectories, we use CAN as the next step for tracking. Tracking in multi-camera environment is challenging owing to variation in lighting conditions, change in human posture, presence of occlusion or blind-spots. To handle all the above challenges in a principled way, we use CAN to weaken the noisy features corresponding to a trajectory and narrow down variances for an identity, thus making the trajectory representation more discriminative.

The input during testing is the set of trajectories $S$ and the output of this step is an association matrix, whose scores are determined using CAN. For every pair of trajectories, appearance feature maps and metadata sequences of one are considered as gallery FV and gallery metadata respectively; while the other forms the probe FV. Probe metadata is appended to each occurrence of gallery metadata. Using EvalNet, we obtain the aggregated gallery FV that corresponds to gallery trajectory's representation. The association score between aggregated FV and probe FV is determined using cosine similarity. In this manner, we perform pairwise trajectory feature similarity comparisons. Final matching/ associations are based on a simple greedy algorithm \cite{narayan2017person} which is equivalent to selecting the largest association score sequentially. This controls proliferation of fragmented identities i.e. a subject is associated with one identity only. 

\section{Experiments and Results}
\label{sec:experiments}
\subsection{DukeMTMC Dataset} 
It consists of more than $2,000$ identities, with over $2$ million frames of $1080$p, $60$fps video, approximately $85$ minutes of video for each of the $8$ cameras \cite{ristani2016performance}. We report results across all $8$ cameras to demonstrate the efficiency of our approach for MPMCT. We use the ground-truth available for the training set (called trainval) of DukeMTMC for evaluation. Only $25\%$ of this set (we call this `net-train') is used for CNN and CAN training and the remaining $75\%$ (`net-test') is used for testing.

\subsection{Comparison with Baseline}
The reference (baseline) approach \cite{ristani2016performance} for DukeMTMC is based on correlation clustering using a binary integer program (BIPCC). The task of tracking is formulated as a graph partitioning problem. Correlations are computed from both appearance descriptors and simple temporal reasoning. 

\subsection{Single-camera Tracking (SCT) Results}
Our system aggregates detection responses into short tracklets using motion and appearance correlation as described in Section~\ref{motionapp}. These tracklets are further aggregated into single camera trajectories using CAN. In Table \ref{comparison1} and Table \ref{comparison2}, SCT performances are reported on each one of the cameras. We compare quantitative performance of our method with the baseline. For a fair comparison, the input to our method are the person detections obtained from MOTchallenge DukeMTMC. The average performance ($IDF1$) over $8$ cameras is $96.58$ using our approach and $89.18$ using BIPCC. An improvement of over $8\%$ is achieved in $IDF1$, approximately $6\%$ in $IDP$ and $10\%$ in $IDR$ metrics. Thus, the results demonstrate that the proposed CAN approach improves within-camera tracking performance. 


\begin{table*}
\begin{center}
\begin{tabular}{|c||c|c|c||c|c|c||c|c|c||c|c|c|}
  \hline
  \multirow{2}{*}{Method} 
      & \multicolumn{3}{c||}{Cam 1} 
          & \multicolumn{3}{c||}{Cam 2}
          	& \multicolumn{3}{c||}{Cam 3}
          		& \multicolumn{3}{c|}{Cam 4}\\             \cline{2-13}
  & $IDP$ & $IDR$ & $IDF_1$ & $IDP$ & $IDR$ & $IDF_1$ & $IDP$ & $IDR$ & $IDF_1$ & $IDP$ & $IDR$ & $IDF_1$\\  \hline
  BIPCC\cite{ristani2016performance} & $92.8$ & $95.8$ & $94.2$ & $93.2$ & $84.8$ & $88.8$ & $94.1$ & $91.0$ & $92.5$ & $88.2$ & $93.4$ & $90.7$\\      \hline
  Ours & 94.5 & 98.3 & \textbf{96.4} & 94.5 & 97.9 & \textbf{96.2} & 98.7 & 97.6 & \textbf{98.1} & 98.9 & 99.6 & \textbf{99.2}\\      \hline
\end{tabular}
\end{center}
\vspace{-0.3cm}
\caption{SCT ID score comparison}
\label{comparison1}
\end{table*}


\begin{table*}
\begin{center}
\begin{tabular}{|c||c|c|c||c|c|c||c|c|c||c|c|c|}
  \hline
  \multirow{2}{*}{Method} 
      & \multicolumn{3}{c||}{Cam 5} 
          & \multicolumn{3}{c||}{Cam 6}
          	& \multicolumn{3}{c||}{Cam 7}
          		& \multicolumn{3}{c|}{Cam 8}\\             \cline{2-13}
  & $IDP$ & $IDR$ & $IDF_1$ & $IDP$ & $IDR$ & $IDF_1$ & $IDP$ & $IDR$ & $IDF_1$ & $IDP$ & $IDR$ & $IDF_1$\\  \hline
  BIPCC\cite{ristani2016performance} & $96.0$ & $81.5$ & $88.1$ & $80.1$ & $72.0$ & $75.9$ & $94.8$ & $91.6$ & $93.2$ & $94.9$ & $85.7$ & $90.1$\\      \hline
  Ours & $95.6$ & $92.2$ & $\textbf{93.8}$ & 96.5 & 90.1 & \textbf{93.2} & 99.3 & 97.1 & \textbf{97.6} & 99.2 & 96.0 & \textbf{98.2}\\      \hline
\end{tabular}
\end{center}
\vspace{-0.3cm}
\caption{SCT ID score comparison}
\label{comparison2}
\end{table*}

\begin{table}
\begin{center}
\begin{tabular}{|c||c|c|c||c|}
  \hline
  \multirow{2}{*}{Method} 
      & \multicolumn{3}{c||}{Identification Score} 
          & \multicolumn{1}{c|}{Our Measure}\\             \cline{2-5}
  & $IDP$ & $IDR$ & $IDF_1$ & $IE (\%) $ \\  \hline
  BIPCC\cite{ristani2016performance} & $76.7$ & $50.0$ & $60.5$ & 1.3 \\      \hline
  Ours & $96.6$ & $58.5$ & $\textbf{72.9}$ & \textbf{0.6}\\      \hline
\end{tabular}
\end{center}
\vspace{-0.2cm}
\caption{ICT (across all $8$ cameras) performance}
\label{comparison3}
\end{table}

\subsection{Inter-camera Tracking (ICT) Results}
We use CAN for ICT. This demonstrates the ability of CAN to scale to SCT as well as ICT. We compare our results with BIPCC using Inference Error rate and Identification scores. Tracking performance across all $8$ cameras is presented in Table \ref{comparison3}. An improvement of over $20\%$ is achieved in $IDF1$, $25\%$ in $IDP$ and $17\%$ in $IDR$ metrics. Thus, the results demonstrate that the proposed CAN approach improves multi-camera tracking performance.
\section{Conclusion}
For representation, instead of using one random feature vector, we can use a set of feature vectors available for a trajectory. Most of the previous methods determine the aggregation weights by only considering the features. Using additional orthogonal information such as metadata corresponding to each feature vector
would lead to discovering better aggregation weights. Using CAN, we can measure the relative quality of every feature map (spatial) in a trajectory for aggregation based on the location and camera information (temporal). It becomes more powerful for video-based person re-identification or multi-camera person tracking by adaptively learning weights and aggregating all detections in the trajectory into a compact feature representation. Moreover, it can be incorporated into any existing re-identification feature representation framework to obtain an optimal template feature for SCT or ICT.


In addition, we study error measures of a tracking system. We define the new measure that emphasizes on tracker mismatches at every timestep. A traditional re-identification system aims at matching person images often on a fixed gallery and cite performance using CMC \cite{martinel2012re} and ROC \cite{hamdoun2008person} curves. However, with a dynamically growing reference set (gallery), it is important to investigate the mismatches. The inference error metric overcomes the disadvantages of existing metrics and provides a better estimate of tracking and re-identification error. We hope this work will benefit researchers working on video surveillance and tracking problems. 


%



\ifCLASSOPTIONcompsoc
  \section*{Acknowledgments}
\else
  \section*{Acknowledgment}
\fi

This material is based upon work supported by the National Science Foundation under Grant IIP \#1266183.





\begin{thebibliography}{1}
\bibitem{babenko2015aggregating}
A.~Babenko and V.~Lempitsky.
\newblock Aggregating local deep features for image retrieval.
\newblock In {\em ICCV}, 2015.

\bibitem{berclaz2011multiple}
J.~Berclaz, F.~Fleuret, E.~Turetken, and P.~Fua.
\newblock Multiple object tracking using k-shortest paths optimization.
\newblock {\em IEEE transactions on pattern analysis and machine intelligence}, 2011.


\bibitem{cai2014exploring}
Y.~Cai and G.~Medioni.
\newblock Exploring context information for inter-camera multiple target
  tracking.
\newblock In {\em Winter
  Conference on Applications of Computer Vision (WACV)}, 2014.

\bibitem{chen2017equalized}
W.~Chen, L.~Cao, X.~Chen, and K.~Huang.
\newblock An equalized global graph model-based approach for multicamera object
  tracking.
\newblock {\em IEEE Transactions on Circuits and Systems for Video Technology}, 2017.



\bibitem{dicle2013way}
C.~Dicle, O.~I. Camps, and M.~Sznaier.
\newblock The way they move: Tracking multiple targets with similar appearance.
\newblock In {\em ICCV}, 2013.

\bibitem{gilbert2006tracking}
A.~Gilbert and R.~Bowden.
\newblock Tracking objects across cameras by incrementally learning
  inter-camera colour calibration and patterns of activity.
\newblock In {\em ECCV}, 2006.

\bibitem{hamdoun2008person}
O.~Hamdoun, F.~Moutarde, B.~Stanciulescu, and B.~Steux.
\newblock Person re-identification in multi-camera system by signature based on
  interest point descriptors collected on short video sequences.
\newblock In {\em 2nd ACM/IEEE International Conference on Distributed Smart
  Cameras (ICDSC)}, 2008.


\bibitem{huang2017densely}
G.~Huang, Z.~Liu, L.~Van Der~Maaten, and K.~Q. Weinberger.
\newblock Densely connected convolutional networks.
\newblock {\em In CVPR}, 2017.

\bibitem{iqbal2016pose}
U.~Iqbal, A.~Milan, and J.~Gall.
\newblock Pose-track: Joint multi-person pose estimation and tracking.
\newblock {\em In CVPR}, 2016.

\bibitem{jin2019multi}
S.~Jin, W.~Liu, W.~Ouyang, and C.~Qian.
\newblock Multi-person Articulated Tracking with Spatial and Temporal Embeddings
\newblock {\em In CVPR}, 2019.

\bibitem{kalal2012tracking}
Z.~Kalal, K.~Mikolajczyk, and J.~Matas.
\newblock Tracking-learning-detection.
\newblock {\em IEEE transactions on pattern analysis and machine intelligence}, 2012.

\bibitem{kviatkovsky2013color}
I.~Kviatkovsky, A.~Adam, and E.~Rivlin.
\newblock Color invariants for person reidentification.
\newblock {\em IEEE Transactions on Pattern Analysis and Machine Intelligence}, 2013.

\bibitem{liao2015person}
S.~Liao, Y.~Hu, X.~Zhu, and S.~Z. Li.
\newblock Person re-identification by local maximal occurrence representation
  and metric learning.
\newblock In {\em CVPR}, 2015.


\bibitem{liu2017quality}
Y.~Liu, J.~Yan, and W.~Ouyang.
\newblock Quality aware network for set to set recognition.
\newblock In CVPR, 2017.


\bibitem{mansfield2002best}
A.~J. Mansfield and J.~L. Wayman.
\newblock Best practices in testing and reporting performance of biometric
  devices.
\newblock CESG, Nat. Phys. Lab., Teddington,
U.K., NPL Tech. Rep. CMSC 14/02, 2002.
\newblock Available: \url{http://www.npl.co.uk/upload/pdf/biometrics_bestprac_v2_1.pdf}

\bibitem{martinel2012re}
N.~Martinel and C.~Micheloni.
\newblock Re-identify people in wide area camera network.
\newblock In {\em Computer Vision and Pattern Recognition Workshops (CVPRW),
  2012 IEEE Computer Society Conference on}, 2012.


\bibitem{nam2016learning}
H.~Nam and B.~Han.
\newblock Learning multi-domain convolutional neural networks for visual
  tracking.
\newblock In {\em Proceedings of the IEEE Conference on Computer Vision and
  Pattern Recognition}, pages 4293--4302, 2016.

\bibitem{narayan2017person}
N.~Narayan, N.~Sankaran, D.~Arpit, K.~Dantu, S.~Setlur, and V.~Govindaraju.
\newblock Person re-identification for improved multi-person multi-camera
  tracking by continuous entity association.
\newblock In {\em Computer Vision and Pattern Recognition Workshops (CVPRW),
  2017 IEEE Conference on}, 2017.

\bibitem{narayan2018re}
N.~Narayan, N.~Sankaran, S.~Setlur, and V.~Govindaraju.
\newblock Re-identification for online person tracking by modeling space-time
  continuum.
\newblock In {\em Proceedings of the IEEE Conference on Computer Vision and
  Pattern Recognition Workshops}, 2018.



\bibitem{ristani2016performance}
E.~Ristani, F.~Solera, R.~Zou, R.~Cucchiara, and C.~Tomasi.
\newblock Performance measures and a data set for multi-target, multi-camera
  tracking.
\newblock In {\em ECCV}, 2016.

\bibitem{ristani2014tracking}
E.~Ristani and C.~Tomasi.
\newblock Tracking multiple people online and in real time.
\newblock In {\em ACCV}, 2014.

\bibitem{sadeghian2017tracking}
A.~Sadeghian, A.~Alahi, and S.~Savarese.
\newblock Tracking the untrackable: Learning to track multiple cues with
  long-term dependencies.
\newblock In ICCV, 2017.

\bibitem{icb17}
N.~Sankaran, S.~Setlur, S.~Tulyakov, and V.~Govindaraju.
\newblock Metadata-based feature aggregation network for face recognition.
\newblock In {\em ICB},
  2017.

\bibitem{wang2015transferring}
N.~Wang, S.~Li, A.~Gupta, and D.-Y. Yeung.
\newblock Transferring rich feature hierarchies for robust visual tracking.
\newblock {\em arXiv preprint arXiv:1501.04587}, 2015.

\bibitem{xu2017jointly}
S.~Xu, Y.~Cheng, K.~Gu, Y.~Yang, S.~Chang, and P.~Zhou.
\newblock Jointly attentive spatial-temporal pooling networks for video-based
  person re-identification.
\newblock In {\em ICCV}, 2017.



\bibitem{yang2017neural}
J.~Yang and P.~Ren.
\newblock Neural aggregation network for video face recognition.
\newblock In CVPR 2017.

\bibitem{yooaction}
S.~Y. J. C.~Y. Yoo, K.~Yun, and J.~Y. Choi.
\newblock Action-decision networks for visual tracking with deep reinforcement
  learning.


\bibitem{zhang2016learning}
L.~Zhang, T.~Xiang, and S.~Gong.
\newblock Learning a discriminative null space for person re-identification.
\newblock In {\em CVPR}, 2016.

\bibitem{zhang2015tracking}
S.~Zhang, Y.~Zhu, and A.~Roy-Chowdhury.
\newblock Tracking multiple interacting targets in a camera network.
\newblock {\em Computer Vision and Image Understanding}, 2015.

\bibitem{ristani2018features}
E.~Ristani, and C.~Tomasi.
\newblock Features for Multi-Target Multi-Camera Tracking and Re-Identification.
\newblock In {\em CVPR}, 2018.

\bibitem{Pernici_2018_CVPR}
F.~Pernici, F.~Bartoli, M.~Bruni, and A.~D.~ Bimbo.
\newblock Memory Based Online Learning of Deep Representations From Video Streams.
\newblock In {\em CVPR}, 2018.

\bibitem{zhu2018online}
J.~Zhu1, H.~Yang1, N.~Liu, M.~Kim, W.~Zhang, and M.~Yang.
\newblock Online multi-object tracking with dual matching attention networks.
\newblock In {\em ECCV}, 2018.

\bibitem{ioffe2015batch}
S.~Ioffe, and C.~Szegedy.
\newblock Batch Normalization: Accelerating Deep Network Training by Reducing Internal Covariate Shift.
\newblock In {\em ICML}, 2015.

\bibitem{glorot2011deep}
X.~Glorot, A.~Bordes, and Y.~Bengio.
\newblock Deep sparse rectifier neural networks.
\newblock In {\em AISTATS}, 2011.

\end{thebibliography}
%
\pagebreak

\end{document}